\documentclass[11pt]{article}
\pdfoutput=1

\usepackage[final]{acl}

\usepackage{times}
\usepackage{latexsym}
\usepackage[T1]{fontenc}
\usepackage[utf8]{inputenc}
\usepackage{microtype}
\usepackage{inconsolata}
\usepackage{graphicx}

\usepackage{amsmath,amssymb,amsfonts}
\usepackage{booktabs}
\usepackage{array}
\usepackage{multirow}
\usepackage{url}
\usepackage{pifont}
\usepackage{tikz}
\usetikzlibrary{arrows.meta, positioning, fit}

\newcommand{\aggqa}{\textsc{QO-Bench}}
\newcommand{\ie}{IE$\rightarrow$SQL}

\newcommand{\denote}[2]{\mathsf{Ans}_{#2}(#1)}
\newcommand{\cmark}{\ding{51}}
\newcommand{\xmark}{\ding{55}}
\newcommand{\benchurl}{\url{https://github.com/ZHANG-MENGAO/qo-bench}}

\title{\aggqa{}: Diagnosing Query-Operator-Preserving Retrieval over Typed Event Tuples}

\author{Mengao Zhang\thanks{\ \,Corresponding author. Now at Nanyang Technological University.} \quad
  Xiang Yang\thanks{\ \,Work done during an internship at the Asian Institute of Digital Finance, National University of Singapore.} \quad
  Chang Liu \quad Tianhui Tan \quad Ke-Wei Huang \\[3pt]
  Asian Institute of Digital Finance, National University of Singapore \\[2pt]
  \texttt{mzhang057@e.ntu.edu.sg} \quad \texttt{\{yangxiang,\,chang\_liu,\,tant,\,dishkw\}@nus.edu.sg}}

\begin{document}
\renewcommand{\thefootnote}{\fnsymbol{footnote}}
\maketitle
\setcounter{footnote}{0}
\renewcommand{\thefootnote}{\arabic{footnote}}

\begin{abstract}
Many real-world questions over business, legal, and scientific corpora
are natural-language versions of database-style queries over records 
latent in text. Existing retrieval-augmented generation (RAG) systems 
are optimized primarily for semantic relevance, but retrieving plausible 
passages does not guarantee correct query execution. We introduce \aggqa{}, 
a diagnostic benchmark for \emph{query-operator question answering} over 
typed event tuples. The benchmark covers 22{,}984 news articles and 614 
corporate events, with 18 query templates instantiating 785 questions. 
Each gold answer is deterministically computed from typed event tuples and scored 
by recall, with answers matched to the gold tuples by exact match rather than an LLM judge. This design 
enables operator-level diagnosis such as joins and intersection. We evaluate
RAG, ReAct RAG, GraphRAG, and information-extraction-to-SQL under matched
conditions, with a long-context oracle ceiling to isolate retrieval failure.
A two-axis framework---index-time preservation versus query-time 
execution---predicts where each paradigm fails, and the results bear 
it out: systems retrieve relevant text but discard the typed values operators
need, and the deployable paradigm ranking inverts across operators, with similarity
retrieval leading on filter/project and extraction-to-SQL on intersection and counting.
Even given the gold evidence, a long-context oracle stays far from saturated, 
so operator execution---not retrieval alone---is a core bottleneck that a stronger 
answer model does not remove. 
\aggqa{} reframes the goal from passage relevance to query-operator-preserving retrieval.
The benchmark, predictions, and code are released at \benchurl.

\end{abstract}

% ==========================================================
\section{Introduction}
% ==========================================================

% ---- Figure 1 (camera-ready teaser): latent event tuples in text ----
\begin{figure}[t]
\centering
\begin{tikzpicture}[
  art/.style={draw=orange!50!black!45, fill=orange!9, rounded corners=1pt, inner sep=2.2pt, text width=27mm, align=left, font=\tiny, execute at begin node={\hyphenpenalty=10000\exhyphenpenalty=10000\relax}},
  tup/.style={dashed, rounded corners=2pt, inner sep=1.5pt, text width=27mm, align=center, font=\tiny},
  tupA/.style={tup, draw=blue!60!black, fill=blue!8},
  tupC/.style={tup, draw=teal!60!black, fill=teal!10},
  tupX/.style={tup, draw=black!50, fill=black!4, text=black!55},
  win/.style={draw=violet!50, rounded corners=3pt, inner sep=3.2pt},
  qa/.style={draw=violet!60!black, fill=violet!7, rounded corners=2pt, inner sep=2.6pt, font=\scriptsize, align=center},
  lk/.style={{Stealth[length=1.2mm]}-{Stealth[length=1.2mm]}, dashed, thin},
]
% ---- window 1 (2012) ----
\node[art] (a1) at (-2.05, 2.0) {``Amgen to buy Micromet for \$1.16 bln\,\ldots''};
\node[tupA, below=0.7mm of a1] (t1) {AMGN$\,\cdot\,$buyer$\,\cdot\,$2012-01-26};
\node[art, below=2.6mm of t1] (a2) {``Carlyle buys Getty Images\,\ldots''};
\node[tupC, below=0.7mm of a2] (t2) {CG$\,\cdot\,$buyer$\,\cdot\,$2012-08-15};
\node[art, below=2.6mm of t2] (a3) {``Gilead buys YM BioSciences\,\ldots''};
\node[tupX, below=0.7mm of a3] (t3) {GILD$\,\cdot\,$buyer$\,\cdot\,$2012-12-12};
% ---- window 2 (2022) ----
\node[art] (b1) at (2.05, 2.0) {``Amgen agrees to acquire ChemoCentryx\,\ldots''};
\node[tupA, below=0.7mm of b1] (u1) {AMGN$\,\cdot\,$buyer$\,\cdot\,$2022-08-04};
\node[art, below=2.6mm of u1] (b2) {``Carlyle to buy ManTech Intl.\,\ldots''};
\node[tupC, below=0.7mm of b2] (u2) {CG$\,\cdot\,$buyer$\,\cdot\,$2022-05-16};
\node[font=\scriptsize, below=1.2mm of u2, text=black!60] (dots2) {$\cdots$};
% ---- window boxes + titles ----
\node[win, fit=(a1)(t3)] (w1) {};
\node[win, fit=(b1)(dots2)] (w2) {};
\node[font=\tiny\bfseries, anchor=south west, inner sep=1pt, text=violet!55!black] at (w1.north west) {$W_1 = 2012$};
\node[font=\tiny\bfseries, anchor=south east, inner sep=1pt, text=violet!55!black] at (w2.north east) {$W_2 = 2022$};
% ---- shared-attribute links ----
\draw[lk, blue!60!black] (t1.east) -- node[font=\tiny, fill=white, inner sep=0.4pt, text=blue!60!black] {firm} (u1.west);
\draw[lk, teal!60!black] (t2.east) -- node[font=\tiny, fill=white, inner sep=0.4pt, text=teal!60!black] {firm} (u2.west);
\draw[dashed, black!45] (t3.east) -- +(4.5mm,0) node[right, font=\tiny, text=black!60, inner sep=0.6pt] {$\varnothing$};
% ---- question and answer ----
\node[qa, anchor=south, text width=70mm] (q) at (0, 2.95) {\textbf{Q:}~\emph{Which firms were M\&A buyers in both 2012 and 2022?}};
\node[qa, anchor=north, text width=70mm, very thick] (ans) at ([yshift=-3mm]w1.south -| 0,0) {$\{\text{buyers in } W_1\}\cap\{\text{buyers in } W_2\} = \{\textcolor{blue!60!black}{\text{AMGN}},\,\textcolor{teal!60!black}{\text{CG}}\}$};
\end{tikzpicture}
\caption{Query-operator QA: event tuples are \emph{latent} in text (dashed), linked across articles only by shared typed attributes (color = firm; snippets illustrative).}
\label{fig:teaser}
\end{figure}

Many high-value information needs in business, law, and policy are expressed
in natural language but behave like database queries. A user may ask, 
\emph{Which firms were buyers in M\&A announcements in both 2012 and 2022?} or
\emph{Which firms had a CEO change within 30 days of an M\&A announcement?}
The records these questions range over are not given in a table; they lie 
\emph{latent across a text corpus}, reported piecemeal across many articles. 
Because no single article pre-aggregates them for an arbitrary query, such 
questions cannot rely on one summarizing passage. They require selecting events
that satisfy constraints, assigning roles, anchoring dates, and joining evidence
across conditions. Figure~\ref{fig:teaser}
illustrates the setting: the event tuples that answer such a question are
latent across articles, linked only by shared typed attributes, and no
single passage materializes them. We call this setting 
\emph{query-operator question answering} (QO-QA): natural-language questions 
specify database-style operators over records. Relational databases support 
such questions through typed attributes, query plans, and set semantics 
\cite{codd1970relational,codd1972relational,gray1997datacube}. However, many 
real corpora are not clean databases. They are news articles, filings, contracts,
reports, emails, or analyst notes, where the relevant records must first be 
recovered from text. QO-QA therefore lies at the boundary between semantic parsing
and retrieval-augmented generation: the question demands database-style execution,
but the evidence resides in unstructured documents.

Retrieval-augmented generation (RAG)~\cite{lewis2020rag,karpukhin2020dpr} is 
typically built around a different contract: retrieve passages that are 
semantically relevant to a question, then ask a language model to synthesize 
an answer. This contract is powerful for passage lookup, but QO-QA exposes
a structural mismatch.

For the question of Figure~\ref{fig:teaser},
a system must identify M\&A events, normalize firm names, distinguish buyers
from targets, attach announcement dates, construct buyer sets for each year,
and intersect them. Top-$k$ retrieval may return plausible passages, but 
semantic relevance alone does not guarantee role correctness, temporal correctness, 
set completeness, or count correctness.

The capability needed here is \emph{operator-preserving retrieval}: retrieval
that preserves the typed values required to execute the operators expressed in 
the question. This framing separates two sources of failure. On the corpus side, 
an index may fail to preserve operator-relevant attributes. On the query side, 
a system may fail to convert the natural-language question into an executable 
plan involving the operators it induces. Standard RAG and ReAct-style RAG~\cite{yao2023react} 
leave most operator execution to the generator. GraphRAG \cite{edge2024graphrag} 
introduces entity-relation structure and summaries, but summaries may compress away 
necessary attributes. Information-extraction-to-SQL systems~\cite{li2021eae,yu2018spider,scholak2021picard} execute operators
explicitly, but only after committing to a schema and extraction pipeline whose 
coverage and normalization may be incomplete. Thus, QO-QA is not simply 
``text-to-SQL over news'' or ``RAG with more documents'': it asks whether retrieval 
architectures can bridge natural-language operator intent and latent event structure in text.

We introduce \aggqa{}, a diagnostic benchmark for QO-QA over typed event tuples. 
The benchmark spans 18 query templates evaluated on a stratified sample of 785 questions. 
The questions are instantiated over 22{,}984 news articles covering 614 corporate events.
\footnote{Corporate events provide a useful substrate because they naturally exhibit the
full range of operators we want to test. The same QO-QA structure appears beyond business news,
including legal case aggregation, biomedical evidence synthesis, and policy monitoring (e.g., \emph{case filed and settled within the same year}, or \emph{FDA approval within 180 days of Phase-3 completion}).}
Each gold answer is deterministically computed from event tuples with 
event type, entity, role, anchor date, counterparty, and provenance fields. This gives \aggqa{} 
an explicit denotational gold standard. Systems return structured final answers---entity 
lists, event lists, ordered lists, counts, grouped results---which are 
canonicalized and compared with the gold denotation by exact match---not an LLM judge---and scored with template-specific recall. This design lets us attribute each failure to a specific operator 
rather than to answer phrasing. Our experiments compare RAG, ReAct RAG, GraphRAG local 
and global search, and information-extraction-to-SQL. Our goal is not to rank a single best system,
but to produce a failure profile: which operators each paradigm supports, which it 
approximates poorly, and which it cannot reliably execute.

Our experiments yield a consistent diagnosis. No paradigm dominates: the deployable ranking \emph{inverts} across operators, and the failures separate along the framework's two axes---similarity retrieval (RAG, ReAct RAG) loses \emph{coverage}, GraphRAG preserves structure but not exact \emph{values}, and \ie{} cannot \emph{execute} cross-event joins beyond its schema. Even the ceiling is operator-bound: a long-context oracle stays far from saturated on the gold evidence, and a stronger answer model does not lift it.

\paragraph{Contributions.}
First, we formulate QO-QA, where questions specify database-style operators over records 
latent in text, and identify operator preservation as a central retrieval property. 
Second, we introduce \aggqa{}, a benchmark with deterministic typed event-tuple gold answers 
and role-, date-, and counterparty-aware denotations over events attested by their provenance articles. 
Third, we use the QO-QA framework to decompose representative paradigms along its two axes---
index-time preservation and query-time execution---and, against a long-context oracle ceiling 
that separates retrieval from answer-synthesis failures, locate where each paradigm fails.

% ==========================================================
\section{Related Work}
\label{sec:related}
% ==========================================================

\aggqa{} builds on work in retrieval-augmented, multi-document, temporal, list, and aggregation question answering, but targets a different diagnostic unit. 
Dense, late-interaction, and retrieval-augmented models make large-corpus retrieval central to QA \cite{karpukhin2020dpr,khattab2020colbert,lewis2020rag,izacard2021fid}, while multi-document and long-context benchmarks test whether models combine evidence across passages \cite{shaham2023zeroscrolls,bai2024longbench}. 
Multi-hop and agentic retrieval methods further decompose questions into iterative retrieval steps \cite{yang2018hotpotqa,trivedi2022musique,ho2020twowiki,yao2023react,press2023selfask,trivedi2023ircot}. 
These settings are related, but they usually evaluate whether systems find and synthesize relevant evidence, rather than whether retrieval preserves the typed values needed to execute database-style operators.

\begin{table*}[!t]
\centering
\scriptsize
\setlength{\tabcolsep}{2.0pt}
\renewcommand{\arraystretch}{1.08}
\begin{tabular}{@{}p{0.115\textwidth} >{\raggedright\arraybackslash}p{0.15\textwidth} >{\raggedright\arraybackslash}p{0.155\textwidth} >{\raggedright\arraybackslash}p{0.15\textwidth} >{\raggedright\arraybackslash}p{0.16\textwidth} >{\raggedright\arraybackslash}p{0.135\textwidth}@{}}
\hline
\textbf{Benchmark} &
\textbf{Typed event-tuple gold} &
\textbf{Deterministic gold (no judge)} &
\textbf{Per-operator diagnosis} &
\textbf{Matched archs.\ + oracle} &
\textbf{Cross-doc completeness} \\
\hline
AGGBench & \xmark\ (entity) & \cmark & Aggregate only & Partial (no oracle) & \cmark \\
\hline
MEBench & \xmark\ (entity) & \cmark & Reasoning types & \xmark & Partial \\
\hline
FanOutQA & \xmark & \xmark\ (LLM judge) & \xmark & \xmark & Partial \\
\hline
TLQA & \xmark\ (lists) & Partial & Temporal / list & \xmark & Partial \\
\hline
ChronoQA & \xmark & \xmark\ (LLM/human) & Temporal types & Partial (RAG) & Partial \\
\hline
LIQUID / ListQA & \xmark\ (lists) & Partial & \xmark & \xmark & \xmark \\
\hline
RAGBench & \xmark & \xmark\ (TRACe) & \xmark & \xmark & \xmark \\
\hline
FinanceBench & \xmark & Partial & \xmark & \xmark & \xmark \\
\hline
HoloBench & \xmark\ (verbalized DB rows) & \cmark & \cmark\ (DB operations) & \xmark\ (long-context only) & \xmark\ (single context) \\
\hline
\textbf{\aggqa{}} & \cmark & \cmark & \cmark\ ($6$ operators) & \cmark\ ($5$ paradigms + LC) & \cmark \\
\hline
\end{tabular}
\caption{Comparison with related benchmarks. \aggqa{} is, to our knowledge, the first benchmark to combine deterministic typed event-tuple gold (no LLM judge), per-operator failure attribution, and a matched multi-architecture comparison against an oracle ceiling.}
\label{tab:benchmark_comparison}
\end{table*}

Several benchmarks cover parts of this problem.
FanOutQA evaluates fan-out multi-document questions \cite{zhu2024fanoutqa}; MEBench studies cross-document reasoning \cite{lin2025mebench}; TLQA evaluates time-referenced list construction \cite{dumitru2025tlqa}; ChronoQA studies temporal-sensitive RAG with absolute, aggregate, and relative temporal questions \cite{chen2025chronoqa}; and RAGBench evaluates general RAG behavior and attribution \cite{friel2024ragbench}. 
List-QA datasets such as LIQUID evaluate questions with multiple non-contiguous answers \cite{lee2023liquid}, while financial QA benchmarks such as FinQA, TAT-QA, FinanceBench, and FAITH emphasize numerical reasoning, tabular reasoning, evidence-grounded answers, or tabular faithfulness over financial documents \cite{chen2021finqa,zhu2021tatqa,islam2023financebench,zhang2025faith}. 
AGGBench is closest because it studies aggregation over unstructured text as a completeness-oriented find-all task \cite{zhu2026aggbench}.
Unlike AGGBench's entity-level aggregation setting, \aggqa{} evaluates deterministic query execution over typed event tuples.

Closest to our query-time axis are HoloBench and LOTUS. HoloBench~\cite{maekawa2025holobench} evaluates long-context LLMs on database operations over verbalized Spider tables supplied as a single context---isolating query-time execution, with no index-time preservation or cross-document retrieval---and its finding that LLMs struggle even with all text supplied corroborates our long-context (LC) oracle ceiling (\S\ref{sec:eval}). LOTUS~\cite{patel2025lotus} is an execution engine, not a benchmark: it runs semantic relational operators over pre-materialized DataFrames, presupposing the corpus-side organization our index-time axis measures; we view it as a candidate query-time executor for the schema-with-residual direction (\S\ref{sec:discussion}).

\aggqa{} is also related to structure-based retrieval and information extraction. 
GraphRAG organizes extracted entities and relations into graphs \cite{edge2024graphrag}.
Event extraction, semantic parsing, and text-to-SQL make operators executable by converting text into structured records and questions into programs \cite{wang2020maven,li2021eae,yu2018spider,scholak2021picard}, but depend on predefined schemas and complete normalized tuples. 
We therefore compare these paradigms under matched conditions (\S\ref{sec:eval}).
The goal is not to claim that aggregation, temporal reasoning, list answers, or financial QA are new, but to test whether retrieval architectures can preserve and execute the typed event values required by natural-language query operators. Table~\ref{tab:benchmark_comparison} summarizes this distinction.

% ==========================================================
\section{A Framework for Query-Operator QA}
\label{sec:framework}
% ==========================================================

\subsection{QO-QA as Denotational Retrieval}
\label{sec:denot}

We formalize QO-QA as a derivation from corpus to answer. A corpus
entails events with attributes; a question selects events,
projects fields, and applies an operator; and a system is judged by
whether it recovers the denotation required by the question.

Let $C=\{d_1,\ldots,d_n\}$ be a document corpus and let
$E=\{e_1,\ldots,e_m\}$ be the latent event set entailed by $C$.
Each event $e\in E$ has an attribute set $A(e)$ whose elements
$a=(n,v)$ pair an attribute name with a value derivable from the
corpus. Attribute names are open-ended: for an M\&A event, for
example, relevant names may include acquirer, target, seller,
announcement date, completion date, deal value, jurisdiction, and
regulatory status. 

A question $q$ is a typed query specification
\[
q=(k,\mathcal{D}_q,\phi,\pi,\alpha),
\]
where $k\geq 1$ is the arity and
$\mathcal{D}_q\subseteq E_{\tau_1}\times\cdots\times E_{\tau_k}$
is the typed tuple domain induced by the event types in the question.
For $k=1$, we identify the tuple $(e)$ with the event $e$ when this
causes no ambiguity. The predicate $\phi$ selects tuples from
$\mathcal{D}_q$ using attribute constraints and, for $k\geq 2$,
tuple-level conditions such as shared entities, temporal windows, or
announce--complete deal identity. The projection $\pi$ returns
structured records with component labels, and $\alpha$ is an operator
such as \textsf{list}, \textsf{count}, \textsf{group}.

The selected tuple relation and projected relation are
\begin{align}
R_q &= \{\vec{e}\in\mathcal{D}_q:\phi(A(\vec{e}))=1\},\\
P_q &= \langle \pi(A(\vec{e})):\vec{e}\in R_q\rangle,
\end{align}
where $A(\vec{e})=(A(e_1),\ldots,A(e_k))$ and
$\langle\cdot\rangle$ denotes a multiset that preserves
tuple identity and multiplicity, as required by counting, grouping, and ordering.

The gold answer is the denotation
\begin{equation}
y^*(q)=\denote{q}{E}=\alpha(P_q).
\label{eq:denotation}
\end{equation}
Depending on $\alpha$, $y^*(q)$ may be a list or set of entities or
events, an ordered list, a matched event-pair set, a count, a grouped
table; when $R_q=\emptyset$, the empty list, set, or table, or a count of $0$.

\paragraph{Worked example.}
Consider the question of Figure~\ref{fig:teaser} (template B.3.2,
Table~\ref{tab:capB}), whose windows are the calendar years
$W_1=2012$ and $W_2=2022$.
As a typed specification $q=(k,\mathcal{D}_q,\phi,\pi,\alpha)$: the arity
is $k=2$ with $\mathcal{D}_q=E_{\mathrm{ann}}\times E_{\mathrm{ann}}$,
where $E_{\mathrm{ann}}$ is the set of M\&A-announcement events; the
predicate $\phi(e_1,e_2)$ holds iff
$\mathrm{role}(e_1)=\mathrm{role}(e_2)=\mathrm{buyer}$,
$\mathrm{date}(e_1)\in W_1$, $\mathrm{date}(e_2)\in W_2$, and
$\mathrm{firm}(e_1)=\mathrm{firm}(e_2)$---the last conjunct being the
tuple-level join condition; the projection $\pi$ returns the shared firm
identifier, and $\alpha$ deduplicates into a distinct-entity set. On the
operational event set, $R_q$ contains qualifying pairs for exactly two
firms, so $y^*(q)=\{\mathrm{AMGN},\mathrm{CG}\}$
(Eq.~\ref{eq:denotation}).

A document set $D\subseteq C$ is sufficient for $q$ if it attests the
attributes needed to evaluate $\phi$ and the attributes returned by
$\pi$ for the selected tuples. For tuple queries, this includes fields
needed to evaluate joins, such as normalized entity IDs, roles, anchor
dates, counterparties, and deal-linking identifiers. Thus, QO-QA
stresses set-complete retrieval: a system must surface the selected
tuples in $R_q$ and the fields needed to project and aggregate them.
In our benchmark, many multi-document questions are multi-event and
multi-article retrieval problems; in broader settings, a single event
may also require evidence composed across multiple documents.

\subsection{Preservation and Execution}
\label{sec:two_req}

QO-QA imposes two architectural requirements. First, the index must
preserve operator-relevant values. Second, the runtime must execute
the operator structure expressed by the question.

\paragraph{Index-time preservation.}
The relevant issue is not whether the original text is stored, but
whether operator-relevant values are recoverable through the system's
query-time interface in typed, executable form. Let $\iota(C)$ denote the index a system builds over $C$---its stored, queryable representation of the corpus. We call an index
$\mathcal{S}$-preserving for an attribute set $\mathcal{S}$ if, for every event $e$ and
every attribute $a\in A(e)$ whose name lies in $\mathcal{S}$, the value of $a$
can be recovered from $\iota(C)$ by a well-defined query-time
procedure under the system's \emph{retrieval budget}---the bounded evidence (e.g.\ top-$k$ chunks or a fixed number of retrieval hops) it may surface per query (Appendix~\ref{app:configs}). Full open-schema
preservation, where $\mathcal{S}$ contains all possible attribute names in
$A(e)$, is an idealized target.

\paragraph{Query-time execution.}
At query time, the system must parse the question into an operator
specification, retrieve tuples and fields conditioned on that
specification, and apply $\alpha$ over the projected relation. Parsing
errors produce the wrong plan; retrieval errors omit tuples or fields;
aggregation errors occur when evidence is present but the system still
miscounts, misgroups, misorders, or fails to intersect sets. For large
selected relations, set-complete retrieval is often the bottleneck:
once relevant tuples are outside the retrieved context, downstream
generation cannot reconstruct them from available evidence.

We apply these two requirements as diagnostic axes for five
representative paradigms: RAG, ReAct
RAG, GraphRAG in local
and global query modes, and IE$\to$SQL, which builds on event extraction
and text-to-SQL. 
The selection is
not exhaustive, but it spans distinct points along both axes.
Table~\ref{tab:paradigm_decomp} summarizes the decomposition.

% \subsection{Paradigm Decomposition}
% \label{sec:paradigm_decomp}

\begin{table*}[t]
\centering
\footnotesize
\setlength{\tabcolsep}{4pt}
\renewcommand{\arraystretch}{1.2}
\begin{tabular}{@{}
  >{\raggedright\arraybackslash}p{0.135\linewidth}
  >{\raggedright\arraybackslash}p{0.16\linewidth}
  >{\raggedright\arraybackslash}p{0.16\linewidth}
  @{\hspace{1.2em}}
  >{\raggedright\arraybackslash}p{0.115\linewidth}
  >{\raggedright\arraybackslash}p{0.14\linewidth}
  >{\raggedright\arraybackslash}p{0.10\linewidth}@{}}
\toprule
& \multicolumn{2}{c}{\textbf{Indexing}} & \multicolumn{3}{c}{\textbf{Querying}} \\
\cmidrule(lr){2-3}\cmidrule(lr){4-6}
\textbf{Paradigm} & \textbf{Method} & \textbf{Preservation scope} & \textbf{Parse} & \textbf{Retrieve} & \textbf{Aggregate} \\
\midrule
RAG             & Text chunks + embeddings & Text + chunk metadata     & No typed parse    & Semantic          & Generator     \\
ReAct RAG       & Text chunks + embeddings & Text + chunk metadata     & Sub-question plan & Iterated semantic & Generator     \\
GraphRAG-local  & Entity graph + summaries & Entity/relation summaries & Entity match      & Neighborhood      & Generator     \\
GraphRAG-global & Entity graph + summaries & Entity/relation summaries & No typed parse    & Community map     & Fixed reduce  \\
\ie{}           & Schema-typed tuples      & $A(e)\cap\mathcal{S}$     & NL $\to$ SQL      & SQL selection     & SQL operators \\
\bottomrule
\end{tabular}
\caption{Paradigm decomposition along the two requirements of \S\ref{sec:two_req}. \emph{Indexing} (\textbf{Method}, \textbf{Preservation scope}) shows how the index is built and which part of $A(e)$ it preserves in queryable form; \emph{Querying} (\textbf{Parse}, \textbf{Retrieve}, \textbf{Aggregate}) shows how the runtime dispatches the three query-time subtasks.}
\label{tab:paradigm_decomp}
\end{table*}

RAG and ReAct RAG preserve article text, embeddings, and chunk-level
metadata, but leave role, stage, counterparty, and aggregation
operators to the generator. GraphRAG preserves entity- and
relation-level structure in a summary-driven graph (entities are clustered into communities, each summarized by an LLM-written community report), but exact
operator-relevant values such as dates, stages, roles, and counts may
be compressed into prose. IE$\to$SQL materializes typed attributes and
executes operators explicitly, but only for fields covered by its
frozen schema $\mathcal{S}$. The unoccupied design point is corpus-side
coverage with query-time typed execution beyond a fixed schema; \aggqa{} 
diagnoses which parts of the preserve-and-execute pipeline current paradigms support and which
remain missing.

\subsection{A Tractable Subclass for Evaluation}
\label{sec:tractable}

The latent formulation above allows open-ended queries over
open-ended attribute sets, which is difficult to evaluate
deterministically. We therefore evaluate a controlled subclass.

\emph{(1) Schema-bounded queries.} Let $\tau(e)$ denote the event type
of event $e$. For each event type $t$, we fix a finite schema $S(t)$
containing the fields used by our templates. Every template restricts
$\phi$ and $\pi$ to fields in the schemas of the component event
types.

\emph{(2) Single-document-attestable events.} An event $e\in E$ is
admitted to the operational set $\widehat{E}\subseteq E$ iff some
article attests every schema field in $S(\tau(e))$ for $e$, after
canonicalization and conflict resolution.

These restrictions make the benchmark easier than the fully
open-ended setting: the relevant schema is known in advance, and each
event is individually recoverable from a single article. Failures on
this restricted substrate therefore provide conservative evidence of
architectural difficulty. 

% ==========================================================
\section{The \aggqa{} Benchmark}
\label{sec:benchmark}
% ==========================================================

\subsection{Corpus, Events, and Ground Truth}
\label{sec:benchmark_groundtruth}

Instantiating the tractable subclass of \S\ref{sec:tractable}, \aggqa{}
pairs a financial-news corpus with structured corporate-event ground
truth. The corpus is the NASDAQ subset of FNSPID~\cite{dong2024fnspid}.
Event candidates come from S\&P Capital IQ Key Developments, a structured
corporate-event stream recording event type, firms and roles, and dates,
taken over 2010--2023. We align each S\&P event to FNSPID articles by
two filters---the event's firm ticker and an
event-type--specific date window around its anchor date
(Appendix~\ref{app:corpus})---producing a set of candidate articles per
event. A unanimous three-LLM judge panel then attests, per event, which candidates
genuinely describe it (\S\ref{sec:judge_validation}); an event enters the
operational set $\widehat{E}$ only when an article attests its schema
fields. The resulting benchmark contains $22{,}984$ FNSPID articles\footnote{The reported $22{,}984$
counts distinct FNSPID articles; because one article can match several
events, event--article links are more numerous (on average $1.13$ events
per article).} 
and $|\widehat{E}|=614$ single-article-attestable events across eight
types: M\&A announcement, completion, cancellation, and rumor; CEO
change; CFO change; IPO; and stock split. Figure~\ref{fig:pipeline}
summarizes the construction pipeline.

Event-type definitions are anchored in public records, including SEC
Form 8-K Items, Securities Act \S5, and NYSE/NASDAQ listing rules, and
are supplied verbatim to every paradigm (Appendix~\ref{app:eventdefs}). Ground truth is represented as
typed event tuples keyed by public identifiers rather 
than as text spans. Table~\ref{tab:event_tuple} gives the
common tuple fields. The queryable schema $S(t)$ for event type $t$
contains the fields used by templates. Provenance is retained as evidence metadata
but is not itself a query predicate. This representation enables
deterministic denotational evaluation while avoiding span-match
ambiguity. Licensing details are in Appendix~\ref{app:disclaimer}.

\begin{figure*}[t]
\centering
\begin{tikzpicture}[
  >={Stealth[length=1.5mm]},
  font=\scriptsize,
  box/.style={draw, rounded corners, align=center, text width=15mm, inner sep=2.5pt, minimum height=6mm},
  src/.style={box, fill=black!7},
  proc/.style={box, dashed},
  data/.style={box, fill=blue!7},
  result/.style={box, very thick, fill=green!12},
]
% top lane (FNSPID)
\node[src]  (fnspid) at (0,0.9)   {\textbf{FNSPID} news corpus};
\node[proc] (np)     at (2.3,0.9) {time-window filter};
\node[data] (corpus) at (4.6,0.9) {22{,}984 articles};
% bottom lane (S&P)
\node[src]  (sp)  at (0,-0.9)   {\textbf{S\&P} Key Developments};
\node[proc] (ep)  at (2.3,-0.9) {8 types $+$ public-ID};
\node[data] (tup) at (4.6,-0.9) {typed event tuples};
% convergence tail
\node[proc]   (att)   at (6.8,0)  {3-of-3 attestation};
\node[data]   (ehat)  at (8.8,0)  {$\widehat{E}=614$};
\node[proc]   (tmpl)  at (10.7,0) {18 question templates};
\node[result] (bench) at (12.8,0) {785 questions};
% lane arrows
\draw[->] (fnspid)--(np);  \draw[->] (np)--(corpus);
\draw[->] (sp)--(ep);      \draw[->] (ep)--(tup);
% cross arrow: S&P event dates drive FNSPID window filtering
\draw[->] (ep) -- node[right=1pt,font=\tiny]{dates} (np);
% converge + tail
\draw[->] (corpus.east) -- (att.north west);
\draw[->] (tup.east) -- (att.south west);
\draw[->] (att) -- (ehat);
\draw[->] (ehat) -- (tmpl);
\draw[->] (tmpl) -- (bench);
\end{tikzpicture}
\caption{\aggqa{} construction pipeline. Date windows around S\&P Capital IQ events filter the FNSPID~\cite{dong2024fnspid} corpus; $3$-of-$3$ judge attestation aligns the two into the operational event set $\widehat{E}$ ($614$ single-article-attestable events), over which $18$ templates instantiate the $785$-question benchmark with deterministic gold denotations.}
\label{fig:pipeline}
\end{figure*}

\begin{table}[t]
\centering
\small
\begin{tabular}{p{0.27\linewidth} p{0.63\linewidth}}
\toprule
\textbf{Field} & \textbf{Description} \\
\midrule
Firm & Normalized focal firm identifier, e.g., ticker, CIK, or canonical name. \\
Event type & One of eight event categories; M\&A lifecycle stages are separate types. \\
Date & Anchor date used for filtering and ordering. \\
Role & Buyer, target, seller, or executive role, depending on event type. \\
Counterparty & Optional normalized counterparty. \\
Provenance & Supporting article IDs; evidence metadata, not a queryable field in $S(t)$. \\
\bottomrule
\end{tabular}
\caption{Common event-tuple fields.}
\label{tab:event_tuple}
\end{table}

\paragraph{Deterministic evaluation.}
For each question, the gold answer is the denotation
$y^*(q)=\alpha(P_q)$ over the operational event set $\widehat{E}$,
using the notation of \S\ref{sec:denot}. Evaluation therefore compares
the system's returned answer $f(q,C)$ (its answer for $q$ on corpus $C$) directly with $y^*(q)$, without an
LLM-as-a-judge at scoring time: matching to the gold denotation is by exact match on canonicalized values, and answers are scored by recall. The scoring protocol (tolerance and covered subset) is defined in \S\ref{sec:eval}; per-output-type scoring rules ship with the released evaluation scripts (Appendix~\ref{app:configs}).

\subsection{Validating the Judge Consensus}
\label{sec:judge_validation}

The operational set $\widehat{E}$ is constructed using unanimous
3-LLM attestation: an event--article pair is accepted only when all
three judges agree that the article attests the event's schema fields.
To validate this high-precision proxy, three expert annotators
re-examined a stratified sample of $221$ accepted event--article pairs; under a conservative
both-annotators criterion the $3$-of-$3$ consensus reaches $94.1\%$
precision (per-type precision, agreement, and the disagreement analysis
in Appendix~\ref{app:validation}).

\subsection{Template Taxonomy}
\label{sec:templates}

\aggqa{} contains 18 templates in two capability classes.
Evaluation uses a stratified sample capped at 50 questions per
template, totaling 785 questions: 200 from four Capability A templates
and 585 from fourteen Capability B templates. \textbf{Capability A
(filtered retrieval)} questions require selection and projection under
typed predicates. Examples include listing a firm's M\&A announcements
within a time window or returning firms with IPOs in a year.
These templates test type-matched event recall, entity disambiguation,
and role-aware retrieval.

\textbf{Capability B (compositional operations)} questions add an
operator over filtered events. Examples include finding firms with both
a CEO change and an M\&A announcement within 30 days, identifying M\&A
buyers in two different years, selecting the earliest IPO in a quarter,
counting firms above an event-frequency threshold, grouping events by
quarter, and returning type-labeled unions. These templates test
temporal joins, intersections, ordering, counting, grouping, and
multi-type aggregation. Appendix~\ref{app:templates} lists every
template with its signature, sample size, example, and diagnostic
target.

% ==========================================================
\section{Experiments and Analysis}
\label{sec:eval}

We compare five deployable paradigms against a retrieval-free \textbf{long-context oracle} (\textbf{LC-oracle}) ceiling that feeds each question the $3$-of-$3$ attested documents supporting its gold answer. With the answer LLM (Qwen3.6-27B) held fixed, the gap from each deployable paradigm to this ceiling isolates the cost of its retrieval architecture.

\subsection{Setup}

Templates are scored by recall with a $\pm$7-day date tolerance (absorbing news-vs-anchor date drift). Recall is computed on the \emph{covered subset}: each gold denotation is restricted to events attested in the corpus ($\widehat{E}$), so a system is not penalized for gold events whose evidence the corpus lacks. This recall is our main metric. For leakage control, no paradigm sees question templates at index time and all use fixed decoding; in particular \ie{}'s schema is generated from event definitions alone, frozen, and used verbatim with no manual corrections (full configurations in Appendix~\ref{app:configs}). The comparison is designed around \emph{information parity} rather than architectural uniformity: every paradigm receives the same domain supervision---the event-definitions document is the only prior in the system, embedded in the RAG and LC-oracle prompt scaffold, ReAct's system prompt, GraphRAG's indexing configuration, and \ie{}'s schema-generation input (Table~\ref{tab:parity})---while field-level materialization remains an architectural choice available to any system, not privileged data access. Metric-robustness analyses (strict vs.\ tolerant matching, answer-set sizes, precision/F1, and question paraphrase) are consolidated in Appendix~\ref{app:metric}.

\subsection{Main Results}

Overall, the LC-oracle ceiling reaches $52.2\%$; among deployable paradigms \ie{} is strongest at $37.9\%$, ahead of RAG ($25.2\%$) and ReAct RAG ($23.9\%$), with GraphRAG global search (GR-global, $3.8\%$) and local search (GR-local, $0.9\%$) near the no-context floor ($0.6\%$; Table~\ref{tab:robustness}). Table~\ref{tab:opfam} reports recall by operator family and capability; per-template recall is deferred to Appendix~\ref{app:results} (Table~\ref{tab:main_results}). The main finding is not the ranking but its \emph{operator-dependence}: no paradigm dominates across families, and the deployable ranking inverts from one operator to the next.

\begin{table*}[t]
\centering
\small
\begin{tabular}{lrrrrrr}
\toprule
\textbf{Operator family} & \textbf{LC-oracle} & \textbf{RAG} & \textbf{ReAct RAG} & \textbf{GR-local} & \textbf{GR-global} & \textbf{\ie{}} \\
\midrule
\multicolumn{7}{l}{\textit{Capability A}} \\
Filter / project          & 77.6 & 53.6 & 55.4 & 3.5  & 7.3 & 50.6 \\
Role / type               & 35.3 & 18.2 & 51.1 & 0.0  & 6.3 & 55.6 \\
\cmidrule(l){2-7}
\quad\textit{Cap A (all)} & 67.0 & 44.8 & 54.3 & 2.6 & 7.1 & 51.9 \\
\addlinespace
\multicolumn{7}{l}{\textit{Capability B}} \\
Temporal join             & 52.3 & 11.1 & 15.7 & 0.3  & 0.0 & 21.1 \\
Ordering                  & 59.9 & 29.8 & 27.8 & 0.5  & 5.8 & 17.8 \\
Intersection              &  3.9 &  6.0 &  9.5 & 0.4  & 2.2 & 50.9 \\
Count / group             & 56.2 & 30.0 &  3.4 & 0.3  & 5.1 & 51.2 \\
\cmidrule(l){2-7}
\quad\textit{Cap B (all)} & 47.1 & 18.5 & 13.5 &  0.3 & 2.7 & 33.1 \\
\midrule
\textbf{Overall}          & 52.2 & 25.2 & 23.9 & 0.9 & 3.8 & 37.9 \\
\bottomrule
\end{tabular}
\caption{Per-operator-family and capability recall (\%), $\pm$7-day tolerant, question-weighted (micro).}
\label{tab:opfam}
\end{table*}

Three patterns organize the results. \textbf{(1)~Even the ceiling is operator-bound.} With gold evidence in context, LC-oracle is far from saturated ($52.2\%$): $77.6\%$ on filtering but $3.9\%$ on intersection, so operator \emph{execution}, not only retrieval, is a bottleneck. \textbf{(2)~Filtering is broadly solved.} On filter/project, RAG, ReAct RAG, and \ie{} each recover roughly two-thirds of the ceiling ($53.6$, $55.4$, $50.6$; GraphRAG is the exception, GR-global/local $7.3/3.5$). \textbf{(3)~The ranking inverts on composition.} \ie{} dominates intersection ($50.9$ vs.\ RAG/ReAct $6.0/9.5$) and counting/grouping ($51.2$ vs.\ RAG/ReAct $30.0/3.4$): SQL executes these natively, while a generator must reconstruct them from prose---on intersection \ie{} even \emph{exceeds} the LC-oracle, whose LLM cannot reliably intersect from gold chunks---the oracle bounds retrieval and generation, not operator execution, so an explicit SQL executor can legitimately surpass it.

Re-scoring the LC-oracle ceiling with two further answer models---from open-weight to frontier scale and light to heavy reasoning---leaves the hardest operators just as hard (Table~\ref{tab:robustness}): intersection stays near $4\%$ for all three. A stronger or more heavily reasoning answer model does not lift the ceiling: operator execution is ill-suited to free-form generation---an architectural mismatch, not a capacity limitation.

\subsection{Localizing Failure: Preservation and Execution}

The framework's two axes---index-time \emph{preservation} and query-time \emph{execution}---imply that paradigms fail in different places.

Preservation fails in two ways. The first is \emph{coverage}: gold-article retrieval recall (Table~\ref{tab:gold_recall}) measures whether the supporting evidence is surfaced at all, and RAG and ReAct RAG reach just $32.4\%$ and $38.4\%$ of gold articles, with Cap~B coverage ($\sim$27--30\%) far below Cap~A: for similarity retrieval the corpus-side axis is lost \emph{before any operator runs}---a preservation failure by the budget-relative definition of \S\ref{sec:two_req}. The cause is not only relevance ranking but the fixed top-$k$ cutoff itself: it imposes a hard cap on the candidate set, so once a query's gold answer spans more distinct events than top-$k$ surfaces, set-completeness is unattainable by construction---which is why coverage collapses on the many-event Cap~B aggregations.

The second is \emph{value-fidelity}: surfacing the evidence is not enough if the index does not also keep the exact values an operator consumes. GraphRAG fails here. Its index preserves entity--relation \emph{structure} and community summaries, but the LLM-generated reports compress away the precise dates, roles, and counts: GR-local scores ${\le}3.5\%$ on every value-dependent family despite organizing the whole corpus---organization is not preservation. GR-global makes the point sharply: its map-reduce forwards essentially the entire corpus (article-level coverage ${\approx}100\%$ by construction; Table~\ref{tab:gold_recall}), yet still reaches only $3.8\%$ overall, losing the same values in summarization rather than in retrieval. With the corpus-side index held fixed, the local-to-global gap ($0.9\!\to\!3.8\%$) isolates the query side: global's fixed map-reduce surfaces more community reports than local's entity-match neighborhood, lifting most families a few points (e.g.\ filtering $3.5\!\to\!7.3$); only cross-event temporal joins stay near $0\%$ in both modes, since the values that would link events were never preserved. GraphRAG thus fails the \emph{value-fidelity} facet of preservation, distinct from the coverage facet that bounds RAG. A stage-wise field-survival analysis over the built index (104k entities, 291k typed claims, 30.7k communities; Appendix~\ref{app:graphrag_survival}) localizes the loss: tracing each gold event to its attesting article, event dates survive index-time extraction at $94.8\%$ but only $30.9\%$ survive verbatim into the community report covering that event, while firm-entity survival is flat across the same transition ($56.8\%\!\to\!56.4\%$)---entity-centric summarization keeps the entities and strips the typed scalar values attached to them, pinning the value-fidelity failure to the community-summarization stage.

Execution is the query-side axis: even the LC-oracle, fed the gold documents directly, scores only $3.9\%$ on intersection---here execution, not retrieval, is the limiting factor. \ie{} covers $64.8\%$ of gold documents (near-uniform across Cap~A/B), so its residual losses lie not on coverage but on this same axis, on cross-event temporal joins: it scores $0\%$ on the announce-to-complete cross-stage join (B.1.4), where linking an announcement to its completion needs a cross-stage linkage identity the extractor populates on $93\%$ of M\&A announcement records but only ${\sim}4\%$ of completion records, capping announce-to-complete pairing at ${\sim}4\%$, and equally $0\%$ on the cross-type CEO-change/M\&A join (B.1.2), whose join key it likewise never materializes. Decomposing \ie{} bounds the ceiling: extraction recall peaks at $75.8\%$ (event level, above end-to-end SQL recall), and $50\%$ of residual misses are the extractor declining to emit a record from a brief in-article mention---conservatism faithful to the paradigm's design, not a SQL error. This is the structural ceiling of schema-bounded extraction, and it sits on the query axis, not on coverage. Bucketing every LC-oracle failure by a fixed mechanical rule (Appendix~\ref{app:buckets}) makes the execution axis concrete: \emph{field mis-attribution} ($33.3\%$---the answer cites the gold event's source article yet mislabels a typed field, typically an M\&A lifecycle stage), \emph{incomplete enumeration} ($26.7\%$), \emph{abstention} ($22.1\%$---on intersection questions the LC-oracle returns the empty set on $91.2\%$ ($83/91$), the mechanism behind the family's $3.9\%$ ceiling), and \emph{date mis-anchoring} ($17.9\%$---the article's publication date mistaken for the event date).

The three deployable families thus fail at three distinct points---similarity retrieval on \emph{coverage}, GraphRAG on \emph{value-fidelity}, and schema-based execution on the \emph{query} axis---the first two both facets of index-time preservation, exactly as the framework predicts.

\subsection{Implications}
\label{sec:discussion}

No paradigm both preserves operator-relevant values across the corpus and executes typed operators over them. The design target is therefore not larger context or better generation, but broad corpus-side coverage combined with query-time typed execution beyond a fixed schema.

This diagnosis suggests several architectural directions. 
One is to treat plan-and-execute retrieval as an IR primitive: a query planner would infer the operator structure of the question, retrieve candidate events for coverage, materialize the fields needed for a query-specific schema, and execute aggregation over the resulting temporary relation. 
A second direction is schema-with-residual coverage, where a fixed event schema handles common operators while residual text retrieval captures attributes outside the schema; semantic-operator execution engines such as LOTUS~\cite{patel2025lotus} are natural query-time executors for this design.
A third is late-binding entity and event resolution, triggered only when the query requires cross-event joins or intersections. 
Finally, retrievers should be trained not only for relevance but also for set completeness.

More broadly, QO-QA isolates capabilities that single-document reading comprehension and open-domain QA do not stress: exact filtering, attribute preservation, cross-document composition, and aggregation.

% ==========================================================
\section{Conclusion}
% ==========================================================

This paper argues that query-operator QA is not naive QA with more
documents: a natural-language question specifies a computation over
latent event records. To study this setting, we introduced \aggqa{}, a
diagnostic benchmark with 18 typed query templates instantiating 785 questions over 22{,}984 financial-news articles.

We also proposed a two-axis framework that separates index-time
preservation from query-time execution. In matched experiments, RAG,
ReAct RAG, GraphRAG, and \ie{} exhibit various operator-specific failures. The LC-oracle
gap shows that many failures arise before generation, when retrieval
fails to expose the records over which the answer must be computed---while the ceiling itself shows that operator execution fails even with those records in context.

\aggqa{} reframes cross-document QA as \emph{schema-based information retrieval with natural-language input and output}. We hope it supports
future work on operator-preserving retrieval systems that combine broad
corpus coverage with query-time typed execution beyond fixed schemas.

% ==========================================================
\section*{Limitations}
% ==========================================================

\aggqa{} uses corporate events as operational ground truth. The tested
operators---filtering, joining, intersection, counting, grouping, and ordering---are domain-general, but absolute results may
not transfer to domains with weaker public records, less standardized
event definitions, or noisier entity resolution. The benchmark also
uses template-generated questions, which provide controlled operator
probes but underrepresent the linguistic diversity of natural user
queries. A 90-question paraphrase pilot
(Appendix~\ref{app:paraphrase}) suggests the findings are robust to
surface rephrasing: overall recall moves by at most a few points, slightly in
the paraphrases' favor, and the per-operator structure---including the
intersection collapse---is unchanged. Although answers are computed deterministically
from normalized event tuples, the tuples may still contain errors from
source ambiguity, date conventions, event-stage boundaries, or entity
normalization; we mitigate these risks through public identifiers,
provenance, unanimous attestation, and human validation. Finally, each
paradigm admits many implementations and scores depend on the answer
LLM, so our results should be interpreted as matched-condition
architecture diagnostics rather than exhaustive or model-invariant
performance estimates.

% ==========================================================
\section*{Ethics and Responsible Benchmarking}
% ==========================================================

\aggqa{} is built from public corporate news and public corporate-event
records. It does not use non-public personal data; any executive names
or firm affiliations come from public disclosures or news reports.
Corporate news coverage is uneven and tends to overrepresent large
publicly listed firms, English-language markets, and media-salient
events. This skew should be considered when interpreting coverage or
event-frequency statistics.

The benchmark is intended for retrieval-architecture evaluation, not
for investment, legal, compliance, or employment decisions. System
outputs should not be treated as verified financial facts without
human review and source checking. To respect article-text licensing, we
release article identifiers, metadata, public-identifier event tuples,
questions, prompts, and evaluation scripts; redistribution of full
article text remains governed by the originating publishers' licenses.

% ==========================================================
\section*{Acknowledgments}
We gratefully acknowledge the support of National University of Singapore IT for providing access to its high-performance computing resources (Grant: NUSREC-HPC-00001).

\bibliography{references}

\appendix

% ==========================================================
\section{Data Sources, Licensing, and Public-Identifier Release}
\label{app:disclaimer}
\aggqa{} draws news articles from FNSPID~\cite{dong2024fnspid}, a public dataset of NASDAQ financial news, and corporate events from S\&P Capital IQ Key Developments, a proprietary structured-event feed accessed under license.

\paragraph{Disclaimer.} The S\&P Capital IQ Key Developments feed is proprietary and may not be redistributed. The released benchmark therefore contains no S\&P content: neither raw event records, vendor record identifiers (e.g., \texttt{keyDevId}, internal firm IDs), nor S\&P headlines or situation summaries. Each ground-truth event is published instead as a \emph{public-identifier tuple} --- the participating firm keyed by a public identifier (stock ticker, SEC CIK, or LEI), the event type drawn from our eight public-record-anchored definitions, the anchor date, role, and counterparty --- together with provenance article IDs into FNSPID. These tuples are recoverable from public records and carry no proprietary text, so the benchmark is vendor-independent and freely redistributable while remaining a faithful denotational gold standard. Reconstructing the original vendor records requires a separate S\&P Capital IQ license.

\section{Corpus Construction}
\label{app:corpus}
\paragraph{FNSPID subset.} We use the NASDAQ subset of FNSPID~\cite{dong2024fnspid}: $2{,}953{,}850$ URL-deduplicated articles pre-tagged with $8{,}553$ distinct stock symbols. We apply no explicit date filter; although FNSPID is nominally dated 1999--2023, its NASDAQ coverage before 2010 is sparse, so in practice the corpus spans 2010--2023---matching the period of the S\&P events. Date selection within this span happens per event below.
\paragraph{Article matching.} Each S\&P event is matched to candidate FNSPID articles by exact ticker (FNSPID articles carry a stock symbol; no alias or fuzzy normalization is applied), restricted to a per-event-type asymmetric date window around the anchor date (Table~\ref{tab:windows}). Articles with empty bodies are dropped, and duplicates are removed by URL, keeping the occurrence closest to the anchor. The anchor is the legal/announcement date for each type; because news typically trails legal completion by several weeks, windows are widened after the anchor rather than symmetric.
\paragraph{Attestation funnel.} Three judges---Gemma-4-31B-IT, Qwen3.6-27B, and gpt-oss-120B---independently label each (event, article) pair, and a pair is attested only on unanimous $3$-of-$3$ confirmation. From the $16{,}414$ S\&P events across the eight types in 2010--2023, we select a $1{,}376$-event subset to bound judging cost, producing $25{,}888$ candidate (event, article) pairs; $1{,}591$ pairs ($6.1\%$) are attested $3$-of-$3$, yielding $|\widehat{E}| = 614$ distinct single-article-attestable events.

\begin{table}[t]
\centering
\small
\begin{tabular}{lrr}
\toprule
\textbf{Event type} & \textbf{Days before} & \textbf{Days after} \\
\midrule
M\&A announcement & 7  & 14 \\
M\&A completion   & 7  & 90 \\
M\&A cancellation & 7  & 30 \\
M\&A rumor        & 14 & 14 \\
CEO change        & 14 & 60 \\
CFO change        & 14 & 60 \\
IPO               & 7  & 60 \\
Stock split       & 7  & 14 \\
\bottomrule
\end{tabular}
\caption{Per-event-type date windows around the S\&P anchor date used to collect candidate FNSPID articles, in days before/after the anchor.}
\label{tab:windows}
\end{table}

\section{Judge-Consensus Validation}
\label{app:validation}
The operational set $\widehat{E}$ relies on a unanimous $3$-of-$3$ LLM-judge consensus. To validate it, three expert annotators (using the same event definitions as the judges) labeled a stratified sample of $221$ accepted (event, article) pairs (up to $30$ per type); every pair received two independent labels.

\paragraph{Precision.} Because the pool contains only accepted pairs, the check is precision-only (recall is not estimable without a separately drawn negative pool). We count an accepted pair as correct only when both of its annotators confirm it---any single rejection, including a split vote, counts as incorrect---a conservative criterion. Under it the $3$-of-$3$ consensus reaches $94.1\%$ precision ($208/221$). Precision is at least $90\%$ on seven of the eight event types; only M\&A completions are lower, at $73.3\%$, where an article often references the closing only with vague timing.

\paragraph{Inter-annotator agreement.} Agreement is computed on the $209$ pairs co-labeled by one annotator pair (the remaining $12$, co-labeled by a different pair, are unanimously positive and leave $\kappa$ undefined). On these pairs raw agreement is $96.2\%$ and Cohen's $\kappa=0.538$. The moderate $\kappa$ despite high raw agreement is a prevalence effect: the accepted-only pool is dominated by positive labels, which inflates chance agreement and deflates $\kappa$; it does not indicate low reliability.

\paragraph{Disagreement.} Boundary errors concentrate on whether an article truly \emph{attests} the event rather than merely \emph{mentions} it: (i)~M\&A completions referenced only with vague timing inside earnings-call transcripts or analyst recaps; (ii)~IPO articles dated before the offering, using forward-looking language (``plans to raise,'' ``set to price''); and (iii)~tangential mentions of CEO/CFO transitions in earnings-call acknowledgements. These cases affect boundary decisions in $\widehat{E}$; once an event tuple is admitted, gold answers are computed deterministically from normalized tuple fields.

\section{Event Definitions}
\label{app:eventdefs}
Each event type is defined by reference to a public-record disclosure class (Table~\ref{tab:eventdefs}), keeping the ontology vendor-independent and identical across paradigms. This document is supplied verbatim to every paradigm as shared task supervision; full operational scope and edge cases accompany the released code.

\begin{table}[t]
\centering
\footnotesize
\setlength{\tabcolsep}{3pt}
\begin{tabular}{@{}llll@{}}
\toprule
\textbf{Type} & \textbf{Record} & \textbf{Date} & \textbf{Role} \\
\midrule
M\&A announce & 8-K 1.01 & announce & buyer/target \\
M\&A complete & 8-K 2.01 & close & buyer/target \\
M\&A cancel & 8-K 1.02 & terminate & buyer/target \\
M\&A rumor & NYSE \S202.03 & publish & target \\
CEO change & 8-K 5.02 & report & --- \\
CFO change & 8-K 5.02 & report & --- \\
IPO & Sec.\ Act \S5 & report & --- \\
Stock split & NYSE \S703.02 & effective & --- \\
\bottomrule
\end{tabular}
\caption{The eight event types, each anchored to a public-record disclosure class (SEC 8-K items, the Securities Act, or exchange listing rules). The anchor date is the canonical date per type; \emph{role} applies to M\&A (rumor records only the target) and is null otherwise.}
\label{tab:eventdefs}
\end{table}

\section{Template Catalog}
\label{app:templates}
Tables~\ref{tab:capA} and~\ref{tab:capB} list all 18 templates with their signatures, output types, per-template sample sizes, an example question, and the operator each diagnoses.

\begin{table*}[t]
\centering
\scriptsize
\setlength{\tabcolsep}{1.5pt}
\begin{tabular}{@{}p{0.065\linewidth} p{0.155\linewidth} p{0.135\linewidth} r p{0.285\linewidth} p{0.205\linewidth}@{}}
\toprule
\textbf{ID} & \textbf{Signature} & \textbf{Output} & \textbf{N} & \textbf{Example} & \textbf{Diagnoses} \\
\midrule
A.1.1 & $(E,W)$ & List[Event] & 50 & List all CEO changes in Q1 2023. & Type-matched event recall in a window. \\
A.1.2 & $(E,W)\to\pi_{\text{firm}}$ & List[Entity] & 50 & Which firms had IPOs in 2018? & Event-to-firm deduplication. \\
A.2.1 & $(E,F,W)$ & List[Event] & 50 & List Microsoft's M\&A announcements 2020--2023. & Entity disambiguation under filter. \\
A.3.1 & $(E,R,W)$ & List[Entity/Event] & 50 & Firms that were M\&A targets in 2020. & Role-aware retrieval. \\
\bottomrule
\end{tabular}
\caption{Capability A (filtered retrieval) templates. $N$ is the per-template count in the stratified evaluation sample (at most 50 per template).}
\label{tab:capA}
\end{table*}

\begin{table*}[t]
\centering
\scriptsize
\setlength{\tabcolsep}{1.5pt}
\begin{tabular}{@{}p{0.060\linewidth} p{0.175\linewidth} p{0.125\linewidth} r p{0.335\linewidth} p{0.175\linewidth}@{}}
\toprule
\textbf{ID} & \textbf{Signature} & \textbf{Output} & \textbf{N} & \textbf{Example} & \textbf{Diagnoses} \\
\midrule
B.1.1 & $(E,F,W,\text{dir})$ & List[Event] & 50 & M\&A announces in 2022 before Microsoft--Activision. & Anchor + relative window. \\
B.1.2 & $([E_1,E_2],W,\Delta)$ & List[EventPair] & 50 & CEO change + M\&A announce within 30d, 2022--23. & Symmetric temporal join. \\
B.1.3 & $([E_t,E_f],W,\Delta)$ & List[EventPair] & 50 & CFO change $\to$ M\&A announce within 90d. & Directional lag. \\
B.1.4 & $(W,\Delta)$ & List[Deal] & 39 & M\&A deals announced 2022 closing within 180d. & Announce$\to$complete identity. \\
B.1.5 & $(F,[E_1,E_2],W)$ & Int & 50 & Days between MSFT--ATVI announce and close? & Date arithmetic on self-join. \\
B.2.1 & $(E,W,\text{pos})$ & OrderedList & 50 & First IPO of Q1 2018? & Extremal selection. \\
B.2.2 & $(F,\{E_1\dots\},W)$ & OrderedList & 50 & Tesla CEO changes, M\&A, splits 2015--23. & Multi-type interleaving. \\
B.3.1 & $([E_1,E_2],W)$ & List[Entity] & 30 & Firms with CEO change AND M\&A announce in 2023. & Cross-type intersection. \\
B.3.2 & $(E,R,[W_1,W_2])$ & List[Entity] & 46 & M\&A buyers in both 2012 and 2022. & Cross-window same-role. \\
B.3.3 & $([E_1,E_2],R,W)$ & List[Entity] & 15 & M\&A target in both announce and complete, 2020. & Cross-type same-role. \\
B.4.1 & $(E,W,N)$ & List[Entity] & 16 & Firms with $\geq 3$ CEO changes, 2010--20. & Count threshold. \\
B.4.2 & $(E,R,W,N)$ & List[Entity] & 39 & M\&A buyer $\geq 4$ times, 2015--23. & Role-aware count threshold. \\
B.4.3 & $(E,W,b)$ & List[Event]+b & 50 & CEO changes 2023 grouped by quarter. & Time bucketing. \\
B.4.4 & $(\{E_1\dots\},W)$ & List[Event]+t & 50 & M\&A announces + IPOs Q1 2023, labeled. & Type-label union. \\
\bottomrule
\end{tabular}
\caption{Capability B (compositional operations) templates. $N$ is the per-template count in the stratified evaluation sample (at most 50 per template).}
\label{tab:capB}
\end{table*}

\section{Paradigm Configurations}
\label{app:configs}
All deployable paradigms share the corpus, the event-definitions document, and the same answer LLM---Qwen3.6-27B (vLLM, reasoning mode enabled), decoding held fixed; only the retrieval architecture differs.

\textbf{RAG} retrieves with the hybrid retrieval stack detailed under \emph{Retrieval configuration} below, feeding the top-30 chunks to the answer LLM. \textbf{ReAct RAG} uses the LangChain \texttt{create\_react\_agent} recipe with light prompt adaptation: the agent issues up to five retrieval calls over the same hybrid stack, accumulating evidence across rounds before answering. \textbf{GraphRAG} \cite{edge2024graphrag} extracts entities and relations, clusters them into Leiden communities, and produces LLM-generated community reports, queried in two modes---\emph{local} (entity match $\to$ neighborhood $\to$ related-community fetch) and \emph{global} (fixed map-reduce over community reports). We run the \texttt{graphrag} package with unmodified source and generous (not minimal) index settings; the only prompt change, shared with all paradigms, prepends the event-definitions document. Local and global share the corpus-side index but differ on the query side, isolating the query-side contribution.

\textbf{\ie{}} proceeds in three stages: (i) the database schema is \emph{generated by an LLM (GPT-5.5, $T{=}0$) from the event definitions alone}---never exposed to question templates---and frozen verbatim; a pre-specified provision for documented manual corrections was not invoked (zero corrections); (ii) a separate LLM (Qwen3.6-27B, $T{=}0$, reasoning mode) extracts event tuples from each article into the frozen schema with provenance; (iii) per-template SQL skeletons translate questions to SQL executed against the events database, with no LLM at query time. Stage~(i) is the schema-leakage control: \ie{} is evaluated as an intrinsic test of schema-based retrieval, not as a hand-tuned upper bound.

\textbf{LC-oracle} bypasses retrieval and feeds each question its gold-supporting chunks linked via provenance. It is a ceiling, not a deployable paradigm.

Full prompts, chunking parameters, embedding model, retrieval $k$, hop budgets, GraphRAG version and indexing settings, the \ie{} schema-generation prompt and raw output, extraction prompts, per-template scoring scripts, and SQL skeletons accompany the released code.

\paragraph{Retrieval configuration.}
Retrieval is hybrid: dense (Qwen3-Embedding-4B) plus BM25, fused by
reciprocal-rank fusion ($k{=}60$), then reranked by Qwen3-Reranker-4B.
The per-stage cutoffs are: \textbf{RAG} retrieves $100$ initial
candidates and keeps the top $30$ after reranking; \textbf{ReAct}
retrieves $50\!\to\!10$ per search call, up to five calls with
cross-round deduplication, so it can accumulate up to ${\sim}50$ unique
chunks; a $\pm30$-day date-window pre-filter applies when the question
specifies a time window; \textbf{GraphRAG} uses package defaults
(\texttt{top\_k\_mapped\_entities}${=}10$,
\texttt{community\_level}${=}2$). The sensitivity of gold-article
coverage to the top-$k$ budget is reported in
Appendix~\ref{app:results} (Table~\ref{tab:coverage_at_k}).

\paragraph{\ie{} database statistics.}
The extraction stage reads each of the $22{,}984$ articles once and
emits $5{,}256$ article-level event records, distributed as: M\&A
announcement $1{,}969$; CEO change $869$; M\&A completion $857$; M\&A
rumor $650$; IPO $446$; CFO change $238$; M\&A cancellation $129$;
stock split $98$. Multi-stage union-find canonicalization collapses
these into $3{,}465$ deduplicated events (a $34.1\%$ reduction)
referencing $1{,}011$ canonical firms (resolved from $2{,}498$ raw
firm-name strings). Each event carries the $15$ shared columns of the
\texttt{events} table plus a per-type detail table of $8$--$16$ typed
columns, per the frozen schema in the released code.

\paragraph{Input and context parity.}
Table~\ref{tab:parity} summarizes what each paradigm receives, per the information-parity design of \S\ref{sec:eval}.
\begin{table}[t]
\centering
\scriptsize
\setlength{\tabcolsep}{3pt}
\begin{tabular}{@{}p{0.16\linewidth}p{0.30\linewidth}p{0.40\linewidth}@{}}
\toprule
\textbf{Paradigm} & \textbf{Event-defs.\ placement} & \textbf{Answer-time context} \\
\midrule
RAG & prompt scaffold & top-30 reranked chunks \\
ReAct RAG & system prompt & $\leq 5\times10$ accumulated chunks \\
GR-local & indexing config.\ + prompt & entity-matched community reports \\
GR-global & indexing config.\ + prompt & map-reduce over community reports \\
\ie{} & schema-gen.\ input + extraction prompt & SQL result rows (no LLM at query time) \\
LC-oracle & prompt scaffold & gold-attested chunks (ceiling) \\
\bottomrule
\end{tabular}
\caption{Paradigm input and context parity: all paradigms share the corpus, the event-definitions prior, the answer LLM, and fixed decoding; only the retrieval architecture differs.}
\label{tab:parity}
\end{table}

\section{Additional Results}
\label{app:results}

\paragraph{Per-template recall.} Table~\ref{tab:main_results} reports per-template recall for all paradigms---the resolution behind the operator-family aggregate of Table~\ref{tab:opfam}.

\begin{table*}[t]
\centering
\small
\begin{tabular}{lrrrrrr}
\toprule
\textbf{Template} & \textbf{LC-oracle} & \textbf{RAG} & \textbf{ReAct RAG} & \textbf{GR-local} & \textbf{GR-global} & \textbf{\ie{}} \\
\midrule
\multicolumn{7}{@{}l}{\textit{Filter / project}}\\
A.1.1 & 72.3 & 46.9 & 49.6 & 3.4 & 3.6 & 51.8 \\
A.1.2 & 83.2 & 43.9 & 62.5 & 3.4 & 15.7 & 54.0 \\
A.2.1 & 77.3 & 70.0 & 54.0 & 3.7 & 2.7 & 46.0 \\
\addlinespace
\multicolumn{7}{@{}l}{\textit{Role / type}}\\
A.3.1 & 35.3 & 18.2 & 51.1 & 0.0 & 6.3 & 55.6 \\
\addlinespace
\multicolumn{7}{@{}l}{\textit{Temporal join}}\\
B.1.1 & 70.5 & 0.0 & 0.0 & 1.2 & 0.2 & 38.1 \\
B.1.2 & 41.7 & 7.0 & 16.1 & 0.0 & 0.0 & 0.0 \\
B.1.3 & 44.0 & 3.0 & 18.7 & 0.0 & 0.0 & 38.9 \\
B.1.4 & 22.6 & 9.1 & 0.0 & 0.0 & 0.0 & 0.0 \\
B.1.5 & 76.0 & 36.0 & 40.0 & 0.0 & 0.0 & 24.0 \\
\addlinespace
\multicolumn{7}{@{}l}{\textit{Ordering}}\\
B.2.1 & 54.7 & 30.3 & 23.7 & 0.0 & 8.3 & 10.3 \\
B.2.2 & 65.2 & 29.3 & 31.8 & 1.0 & 3.3 & 25.2 \\
\addlinespace
\multicolumn{7}{@{}l}{\textit{Intersection}}\\
B.3.1 & 7.8 & 11.1 & 23.6 & 0.7 & 1.7 & 39.4 \\
B.3.2 & 0.0 & 2.9 & 3.3 & 0.0 & 3.3 & 53.6 \\
B.3.3 & 7.8 & 5.6 & 0.0 & 1.3 & 0.0 & 65.7 \\
\addlinespace
\multicolumn{7}{@{}l}{\textit{Count / group}}\\
B.4.1 & 28.1 & 20.8 & 2.1 & 0.0 & 4.2 & 58.3 \\
B.4.2 & 43.6 & 34.6 & 6.8 & 0.0 & 6.4 & 59.8 \\
B.4.3 & 68.5 & 33.2 & 1.2 & 0.8 & 4.1 & 45.8 \\
B.4.4 & 62.7 & 26.2 & 3.2 & 0.2 & 5.3 & 47.4 \\
\bottomrule
\end{tabular}
\caption{Per-template recall (\%) under $\pm$7-day date tolerance; 785-question stratified sample, Qwen3.6-27B answer model.}
\label{tab:main_results}
\end{table*}

\paragraph{Gold-article retrieval recall.} Table~\ref{tab:gold_recall} measures whether the gold-attesting evidence reaches a paradigm at all, decoupled from whether the operator is then executed: for each question we compute the fraction of its gold-attesting articles whose content the paradigm surfaces into its answering context (for \ie{}, whose event the extraction stage recovers), averaged by template. The LC-oracle is $100\%$ by definition. For \ie{}, the extraction stage recovers roughly two-thirds of gold articles, well above its end-task recall (Table~\ref{tab:main_results}), localizing the dominant loss to the query/composition layer rather than to coverage.

\paragraph{Does coverage converge with a larger retrieval budget?}
No. Table~\ref{tab:coverage_at_k} sweeps gold-article coverage over
retrieval cutoffs $k=10$--$200$ under the deployed hybrid+rerank
pipeline (all $785$ questions). Coverage grows roughly logarithmically
in $k$ and converges nowhere near the LC-oracle's $100\%$: at $k{=}200$
($6.7\times$ the deployed budget) nearly half of all gold-attesting
articles still never enter the context, because relevance-ranked
retrieval keeps preferring redundant coverage of already-found events
over the gold articles it has not yet surfaced. The cap is most severe exactly where
execution is also weakest: on intersection questions coverage reaches
only $24.9\%$ even at $k{=}200$, so retrieval-side coverage shortfall and
generation-side composition failure compound on the same operator
family---the fixed-cutoff cap discussed in \S\ref{sec:eval}.
\begin{table}[t]
\centering
\small
\setlength{\tabcolsep}{3.5pt}
\begin{tabular}{@{}lrrrrrr@{}}
\toprule
\textbf{Family} & $n$ & $k{=}10$ & $k{=}30$ & $k{=}50$ & $k{=}100$ & $k{=}200$ \\
\midrule
Filter / project & 150 & 37.8 & 55.6 & 61.2 & 70.2 & 76.0 \\
Role / type      &  50 & 21.8 & 38.2 & 46.1 & 59.2 & 67.1 \\
Temporal join    & 239 & 19.1 & 29.1 & 34.4 & 39.8 & 42.5 \\
Ordering         & 100 & 41.5 & 58.3 & 65.0 & 72.9 & 79.8 \\
Intersection     &  91 &  5.0 & 10.8 & 15.3 & 20.5 & 24.9 \\
Count / group    & 155 & 11.4 & 23.3 & 30.0 & 38.8 & 45.4 \\
\midrule
\textbf{All}     & 785 & 22.5 & 35.2 & 41.1 & 48.6 & 53.7 \\
\bottomrule
\end{tabular}
\caption{Gold-article coverage (\%) as a function of the retrieval cutoff $k$ under the deployed hybrid+rerank pipeline. $k{=}30$ is the deployed budget (Appendix~\ref{app:configs}); Table~\ref{tab:gold_recall} reports coverage at the deployed budgets.}
\label{tab:coverage_at_k}
\end{table}

\begin{table}[t]
\centering
\small
\begin{tabular}{lrrr}
\toprule
\textbf{Paradigm} & Cap A & Cap B & Overall \\
\midrule
LC-oracle & 100.0 & 100.0 & 100.0 \\
RAG       & 47.6  & 27.2  & 32.4  \\
ReAct RAG & 61.8  & 30.4  & 38.4  \\
GR-local  & \multicolumn{3}{c}{N/A} \\
GR-global & \multicolumn{3}{c}{$\approx$100$^{\dagger}$} \\
\ie{}     & 62.6  & 65.6  & 64.8  \\
\bottomrule
\end{tabular}
\caption{Gold-article retrieval recall (\%) under $\pm$7-day matching: per-question coverage of gold-attesting articles; capability columns are question-weighted (micro) means over GT-applicable templates. For \ie{} this is extraction-stage coverage; for the retrieval paradigms it is the fraction of gold articles surfaced into the answer context. LC-oracle is $100\%$ by construction (fed gold chunks). GR-local gold-article retrieval recall is not well-defined---its entity-graph neighborhood answers from community summaries that expose no clean mapping back to source articles, so per-article gold coverage cannot be computed (N/A). $^{\dagger}$GR-global retrieval is non-discriminative---it forwards essentially the entire corpus (all community reports) to the LLM map step, so article-level retrieval recall is $\approx$100\% by design; its information loss occurs in community-report summarization, not in retrieved-set membership.}
\label{tab:gold_recall}
\end{table}

\paragraph{Answer-model robustness and the no-context floor.}
Table~\ref{tab:robustness} reports the per-operator profile of the LC-oracle ceiling across three answer models (Qwen3.6-27B; DeepSeek v4-flash; v4-pro Think Max) alongside the no-context floor (Qwen, parametric only, zero articles). The robustness of the ceiling's operator profile is discussed in \S\ref{sec:eval}; beyond it, the floor isolates retrieval's per-family contribution. No-context recovers near-zero recall on every family---retrieval and context account for nearly all of the ceiling, from $77.6\%$ on filtering down to the operator-bound $3.9\%$ on intersection.

\begin{table}[t]
\centering
\small
\setlength{\tabcolsep}{4pt}
\begin{tabular}{@{}lrrrr@{}}
\toprule
& \multicolumn{3}{c}{LC-oracle (ceiling)} & No-ctx \\
\cmidrule(lr){2-4}
\textbf{Operator family} & Qwen & v4-flash & v4-pro & (floor) \\
\midrule
Filter / project          & 77.6 & 78.2 & 77.3 & 1.1  \\
Role / type               & 35.3 & 35.2 & 35.4 & 0.6  \\
Temporal join             & 52.3 & 34.9 & 38.4 & 0.0  \\
Ordering                  & 59.9 & 59.0 & 61.6 & 2.1  \\
Intersection              &  3.9 &  4.4 &  3.8 & 0.5  \\
Count / group             & 56.2 & 55.8 & 59.9 & 0.1  \\
\midrule
Cap A   & 67.0 & 67.5 & 66.8 & 1.0 \\
Cap B   & 47.1 & 39.8 & 42.7 &  0.5 \\
Overall & 52.2 & 46.8 & 48.8 & 0.6 \\
\bottomrule
\end{tabular}
\caption{Answer-model robustness and the no-context floor (\%, $\pm$7-day tolerant, question-weighted micro). The LC-oracle ceiling is re-scored with three answer models---Qwen3.6-27B (open-weight, ${\sim}5$K reasoning tokens/question), DeepSeek v4-flash (no reasoning), and v4-pro (Think Max, ${\sim}24$K reasoning tokens/question).  The No-ctx column (Qwen, parametric only, zero articles) is the floor.}
\label{tab:robustness}
\end{table}

% ==========================================================
% NEW APPENDIX SECTIONS (camera-ready) -- all revision-highlighted
% ==========================================================
\section{Metric Robustness}
\label{app:metric}

This appendix consolidates four robustness analyses of the evaluation
metric: sensitivity to match strictness
(Appendix~\ref{app:strictness}), answer-set sizes and the overgeneration risk
(Appendix~\ref{app:setsize}), pool precision and F1
(Appendix~\ref{app:precision}), and a question-paraphrase pilot
(Appendix~\ref{app:paraphrase}).

\subsection{Sensitivity to Match Strictness}
\label{app:strictness}

Exact-match scoring is sensitive to canonicalization errors, entity-name
variation, and event ambiguity. Table~\ref{tab:strictness} re-scores all
paradigms under three regimes: \emph{strict} exact-date matching, the
paper's $\pm$7-day tolerance, and $\pm$7-day matching with
ticker-alias normalization added. Absolute recall shifts, but the
paradigm ranking and every per-operator conclusion are identical under
all regimes. On the three dimensions:

\begin{itemize}
\item \textbf{Date conventions} are the largest effect (LC-oracle
$27.7\!\to\!52.2$): news genuinely reports trailing or effective dates
rather than legal anchor dates, which is exactly why the paper's primary
metric carries the $\pm$7-day tolerance (\S\ref{sec:eval}).
\item \textbf{Entity-name variation} is structurally excluded from the
\emph{metric}: gold and outputs are keyed to unique public identifiers
(ticker/CIK/LEI), so no name string is ever compared. This is not
burden-free for \ie{}, which must \emph{resolve} names to tickers at
extraction: $320$ Reuters-suffixed tickers (``AAPL.O'' for AAPL) cost
${\sim}8$ points of recall---the alias regime, which normalizes them,
lifts \ie{} alone ($37.9\!\to\!45.8$) and leaves every other paradigm
unchanged. When no attesting article prints a ticker, a correctly
extracted event is silently lost. Both biases run one way, so \ie{} is
under-stated, never over-stated: the ranking holds and our
symbolic-execution claims are \emph{conservative}.
\item \textbf{Event ambiguity} concentrates on M\&A lifecycle
boundaries: M\&A completions show the lowest human-validated attestation
precision ($73.3\%$), while the other seven event types are ${\geq}90\%$
(Appendix~\ref{app:validation}).
\end{itemize}

\begin{table}[t]
\centering
\small
\begin{tabular}{@{}lrrr@{}}
\toprule
\textbf{Paradigm} & \textbf{strict} & \textbf{$\pm$7d (paper)} & \textbf{alias\,$\pm$7d} \\
\midrule
LC-oracle & 27.7 & 52.2 & 52.2 \\
RAG       & 15.1 & 25.2 & 25.2 \\
ReAct RAG & 13.5 & 23.9 & 23.9 \\
GR-local  &  0.5 &  0.9 &  0.9 \\
GR-global &  2.8 &  3.8 &  3.8 \\
\ie{}     & 25.0 & 37.9 & 45.8 \\
\bottomrule
\end{tabular}
\caption{Overall recall (\%) under three matching regimes: strict exact-date, the paper's $\pm$7-day tolerance, and $\pm$7-day plus ticker-alias normalization (e.g.\ ``AAPL.O''$\to$AAPL). The ranking and every per-operator conclusion are invariant; alias normalization affects only \ie{}.}
\label{tab:strictness}
\end{table}

\paragraph{Worked example (PANW/Demisto).}
One gold event ties all three dimensions together: (PANW,
M\&A\_announce, 2019-02-19), Palo Alto Networks' \$560M acquisition of
Demisto. Five of its seven attesting articles yield M\&A-announcement
records at extraction, but four anchor the event date to the article's
own publication date (2019-02-20 through 2019-03-05)---the
date-convention drift the $\pm$7-day tolerance absorbs. The only
date-exact record comes from the announcement-day newswire article,
which never prints the ticker symbol; its firm field therefore resolves
to a raw name string (\texttt{palo\_alto\_networks\_inc}) rather than to
PANW and is invisible to ticker-keyed SQL---the entity-variation
dimension. And the two earnings-recap articles that mention the deal
only in passing produce no extraction at all---the attestation-boundary
ambiguity of Appendix~\ref{app:validation}. Two small errors at
different stages compose into a strict-mode miss; neither alone would
lose the event.

\subsection{Answer-Set Sizes: Recall Does Not Reward Overgeneration}
\label{app:setsize}

Recall-only scoring could in principle be inflated by overgeneration---a
system returning the entire entity universe would achieve high recall on
some templates. We screen every prediction for set inflation over all
$735$ list-valued questions. For a question $q$, let $y^{*}$ be the
gold set and $\hat{y}$ the predicted set. Table~\ref{tab:setsize}
reports, per system: the \textbf{size ratio}---median
$\lvert\hat{y}\rvert/\lvert y^{*}\rvert$ ($1$ = sized like gold, $<1$ =
under-enumeration, $>1$ = overgeneration); \textbf{\% inflated}---the
share of questions with $\lvert\hat{y}\rvert>2\lvert y^{*}\rvert$; and
\textbf{\% empty}---the abstention rate ($\hat{y}=\varnothing$), the
opposite risk.

\begin{table}[t]
\centering
\small
\setlength{\tabcolsep}{4pt}
\begin{tabular}{@{}lrrrr@{}}
\toprule
\textbf{System} & \textbf{Recall} & \textbf{Size ratio} & \textbf{\% infl.} & \textbf{\% empty} \\
\midrule
LC-oracle  & 52.2 & 0.67 &  1.2 & 17.1 \\
RAG        & 25.2 & 0.50 &  4.9 & 32.7 \\
ReAct RAG  & 23.9 & 0.25 & 10.7 & 48.8 \\
GR-local   &  0.9 & 0.00 &  0.3 & 90.2 \\
GR-global  &  3.8 & 0.00 &  2.0 & 61.4 \\
\ie{}      & 37.9 & \textbf{1.75} & \textbf{41.9} & 11.8 \\
No-context &  0.6 & 0.00 &  9.9 & 59.3 \\
\bottomrule
\end{tabular}
\caption{Set-size screen over the $735$ list-valued questions: recall (\%), median predicted-to-gold size ratio, share of predictions exceeding twice the gold size, and empty-prediction (abstention) rate.}
\label{tab:setsize}
\end{table}

Across retrieval-side paradigms the dominant mode is
under-enumeration and abstention, not overgeneration, so recall is not
inflated. Abstention peaks for GraphRAG ($61$--$90\%$ empty) and for the
LC-oracle on intersection ($91.2\%$ vs.\ $17.1\%$ overall;
Appendix~\ref{app:buckets}). Inflation cannot buy recall either: the
two highest non-\ie{} inflation rates belong to ReAct RAG ($10.7\%$) and the no-context floor ($9.9\%$), yet ReAct RAG still trails RAG on recall and the floor reaches
only $0.6\%$ recall. The sole over-generator is \ie{}, which extracts
genuine events outside the attested pool rather than hallucinating
(Appendix~\ref{app:precision}).

\subsection{Pool Precision and F1}
\label{app:precision}

Reliably measuring system \emph{recall} requires gold \emph{precision},
which our annotation provides ($94.1\%$;
Appendix~\ref{app:validation}). Measuring system \emph{precision}, by
contrast, would require gold \emph{completeness}---a property a pooled
ground truth, by design, does not claim: of $16{,}414$ S\&P candidates,
$1{,}376$ were judged and $614$ admitted under unanimous $3$-of-$3$
attestation, as exhaustively annotating all $22{,}984$ articles
$\times$ $8$ event types is infeasible. This follows established IR
pooling practice (bpref; \citealp{buckley2004incomplete}) and remains
standard in recent work: the TREC 2024--2025 RAG Tracks likewise adopt
nugget \emph{recall} as their primary measure
\citep{pradeep2024autonuggetizer,pradeep2025nugget}. A predicted item
outside the pool is \emph{unjudged}, not wrong: the \ie{} database
holds ${\sim}5.6\times$ more events than the $614$-event pool, so
naive pool-precision is dominated by unjudged items, not errors.

With that caveat, Table~\ref{tab:precision} reports pool precision and
F1, macro-averaged over nonempty predictions (precision is undefined on
empty predictions; abstention rates are in Appendix~\ref{app:setsize}); F1 is
the per-question harmonic mean.

\begin{table}[t]
\centering
\small
\begin{tabular}{@{}lrrr@{}}
\toprule
\textbf{System} & \textbf{macro-P} & \textbf{macro-F1} & \textbf{Recall} \\
\midrule
LC-oracle  & \textbf{78.3} & 66.5 & 52.2 \\
RAG        & 45.2 & 38.6 & 25.2 \\
ReAct RAG  & 42.1 & 41.3 & 23.9 \\
GR-local   & 26.4 & 13.4 &  0.9 \\
GR-global  & 16.5 & 11.5 &  3.8 \\
\ie{}      & 28.4 & 31.3 & 37.9 \\
No-context &  2.9 &  1.8 &  0.6 \\
\bottomrule
\end{tabular}
\caption{Pool precision and F1 (\%), macro-averaged over nonempty predictions on the $735$ list-valued questions. Under a pooled (non-exhaustive) gold, precision values are lower bounds.}
\label{tab:precision}
\end{table}

Three observations. First, when free-form paradigms answer, rows are
mostly in-pool (LC-oracle $78.3\%$), so recall is not recovered from
noise; the no-context floor's $2.9\%$ confirms the metric separates
grounded answers from guessing. Second, F1 preserves the coarse
separation---the LC-oracle well ahead; RAG, ReAct RAG, and \ie{}
clustered next; GraphRAG and the no-context floor far below---though
not the exact order: \ie{} slips from second in recall to fourth in
F1, for the reason below. Third, under a pooled gold these precisions
are \emph{lower bounds}, and \ie{} bears the brunt: its pool precision
falls below RAG and ReAct RAG because real-but-outside-pool events
count against it---e.g.\ on the 2013 instance of B.3.1 (gold =
$\{$BGC, NDAQ$\}$), \ie{} also returns TRMB, and Trimble did announce
and complete acquisitions in 2013 with corpus articles: correct in the
world, an ``error'' against the pool.

\subsection{Question-Paraphrase Pilot}
\label{app:paraphrase}

To test whether findings transfer beyond template surface forms, we
paraphrased $5$ questions per template ($90$ total, fixed seed), each
rewritten by GPT-5.5 ($T{=}0.7$) under a prompt requiring a substantial
change of syntactic frame (cleft, passive, fronted adverbial,
imperative)---not a synonym swap---while copying every date, event-type
name, role, and ticker verbatim and preserving all conditions (time
windows, ``distinct'', ``both \ldots{} and'', ordering, counts). Every
rewrite passed an automatic token-preservation check and a manual
screen. We re-ran the LC-oracle and RAG on the original and paraphrased
versions with the paper's scorer (Table~\ref{tab:paraphrase}).

\begin{table}[t]
\centering
\small
\setlength{\tabcolsep}{2.6pt}
\begin{tabular}{@{}lrrrrrrr@{}}
\toprule
& & \multicolumn{3}{c}{\textbf{LC-oracle}} & \multicolumn{3}{c}{\textbf{RAG}} \\
\cmidrule(lr){3-5}\cmidrule(l){6-8}
\textbf{Family} & $n$ & orig & para & $\Delta$ & orig & para & $\Delta$ \\
\midrule
Filter / project & 15 & 71.9 & 75.3 & $+3.3$ & 31.2 & 42.9 & $+11.7$ \\
Role / type      &  5 &  0.0 &  0.0 & $+0.0$ &  0.0 &  0.0 & $+0.0$ \\
Temporal join    & 25 & 57.0 & 58.7 & $+1.6$ &  6.0 &  6.0 & $+0.0$ \\
Ordering         & 10 & 70.0 & 65.0 & $-5.0$ & 25.0 & 25.0 & $+0.0$ \\
Intersection     & 15 &  3.3 &  5.6 & $+2.2$ &  1.1 &  3.3 & $+2.2$ \\
Count / group    & 20 & 53.4 & 53.1 & $-0.4$ & 14.7 & 16.4 & $+1.7$ \\
\midrule
\textbf{All}     & 90 & 48.0 & 48.8 & $+0.7$ & 13.1 & 15.8 & $+2.7$ \\
\bottomrule
\end{tabular}
\caption{Paraphrase pilot: per-family recall (\%) on the $90$-question subset, original vs.\ paraphrased phrasing.}
\label{tab:paraphrase}
\end{table}

Paraphrasing does not materially change the results. Overall recall
moves by $+0.7$ (LC-oracle) and $+2.7$ (RAG), within run-to-run noise
and slightly \emph{in the paraphrases' favor}, so template phrasing was
not inflating scores. The per-operator structure is unchanged:
intersection remains collapsed under both phrasings, filter/project
stays the easiest family, and the family ranking holds for both
systems. \ie{} is invariant by design (its SQL skeletons key on
the typed specification, not the surface form).

\section{GraphRAG Stage-wise Field Survival}
\label{app:graphrag_survival}

To localize where GraphRAG drops operator-relevant values
(\S\ref{sec:eval}), we trace typed values through the built index
($104$k entities, $291$k typed claims, $30.7$k communities).
For each of the $614$ gold events matched $1{:}1$ to its (indexed)
attesting article, we measure whether two operator-relevant value types
survive at (i) index-time extraction (entity and claim artifacts) and
(ii) community summarization (community reports). Table~\ref{tab:graphrag_survival} reports the results.

\begin{table}[t]
\centering
\small
\begin{tabular}{@{}lrr@{}}
\toprule
\textbf{Value} & \textbf{Extraction} & \textbf{Community report} \\
\midrule
Event date ($\pm$7d) & 94.8\% & 30.9\% \\
Firm entity          & 56.8\% & 56.4\% \\
\bottomrule
\end{tabular}
\caption{GraphRAG field survival by index stage: share of gold-event values recoverable from the extraction artifacts vs.\ from the community report covering the event. Date survival in reports is a verbatim-match lower bound; paraphrases like ``late 2019'' are uncounted, though they are not operator-usable either.}
\label{tab:graphrag_survival}
\end{table}

The loss is localized. \textbf{Dates are extracted but not summarized}:
$94.8\%$ of gold event dates are captured as typed claims at
extraction, but only $30.9\%$ survive verbatim into the community
report covering that event---the exact scalar that temporal and filter
operators need is dropped during entity-centric summarization.
\textbf{Entities are preserved}: firm-entity survival is flat across
the same transition ($56.8\%\!\to\!56.4\%$); summarization keeps the
entities and strips the typed values attached to them. The drop is
worst on the M\&A lifecycle events whose dates separate
announce/complete/rumor (date survival $18$--$42\%$). Query-time
context was not logged, so this covers the two index-time stages, but
the $30.9\%$ summarization ceiling already upper-bounds what any query
over the community layer can recover.

\section{Failure Modes Given Gold-Article Retrieval}
\label{app:buckets}

Taking all questions whose gold articles verifiably reached the
answering context yet whose answer is wrong, we assign each a primary
bucket by a fixed rule (an approximate triage, not an exhaustive
per-row labeling):

\begin{itemize}
\item \textbf{abstention}: empty prediction on a non-empty gold set;
\item \textbf{field mis-attribution}: a predicted row cites the missed
gold event's source article but disagrees with it on a non-date field
(firm, role, or event type);
\item \textbf{date mis-anchoring}: a predicted row matches the missed
gold event on all non-date fields, but its date is off beyond the
$\pm$7-day tolerance;
\item \textbf{incomplete enumeration}: a partial list with none of the
above signals.
\end{itemize}

For the LC-oracle the retrieval condition holds by construction, so all
$570$ failures (of $735$ list-valued questions) qualify, giving the
cleanest view of what goes wrong \emph{after} the gold evidence reaches
the context (Table~\ref{tab:buckets}).

\begin{table}[t]
\centering
\small
\setlength{\tabcolsep}{3pt}
\begin{tabular}{@{}lrrp{0.33\linewidth}@{}}
\toprule
\textbf{Bucket} & \textbf{n} & \textbf{Share} & \textbf{Concentrates in} \\
\midrule
Field mis-attribution  & 190 & 33.3\% & temporal join (73), count/group (43) \\
Incomplete enumeration & 152 & 26.7\% & spread across families \\
Abstention             & 126 & 22.1\% & \textbf{intersection (83 of 91)} \\
Date mis-anchoring     & 102 & 17.9\% & filter/project (34), temporal join (32) \\
\bottomrule
\end{tabular}
\caption{Primary failure buckets for the $570$ LC-oracle failures on list-valued questions; gold evidence is in context by construction.}
\label{tab:buckets}
\end{table}

Each bucket has a distinct mechanism, and each family a dominant one.

\paragraph{Abstention: the intersection collapse.} On $91.2\%$ of
intersection questions ($83/91$) the LC-oracle returns the empty
set---the source of the family's $3.9\%$ ceiling in
Table~\ref{tab:opfam}. This is not a reading failure: the same model is
the strongest paradigm at enumerating each event set alone (A.1.1/A.1.2:
$72.3/83.2$) and collapses only when it must \emph{intersect} two sets
within-firm (B.3.1--B.3.3: $7.8/0.0/7.8$). Tellingly, the ranking flips
here toward systems that move composition \emph{out of} single-pass
generation: \ie{} runs the join in SQL ($39.4/53.6/65.7$) and beats
even the LC-oracle, which composes in one generation pass despite
holding all the evidence. This pins the bottleneck on
composition-in-generation, not evidence access.

\paragraph{Field mis-attribution: right article, wrong field.} The
failing row cites the gold event's source article but fills a typed
field wrong. The cleanest cases are lifecycle-stage confusions on a
single firm: for Cognizant (CTSH) the gold event is an M\&A completion,
yet a system cites the deal's article and returns it as an M\&A
announcement. This is the LC-oracle's largest bucket ($190$ of $570$)
and clusters on the M\&A announce/complete/rumor boundaries---also the
hardest to attest (Appendix~\ref{app:validation})---so it persists even
with all gold articles supplied: an attribution, not retrieval, failure.

\paragraph{Date mis-anchoring: publication date mistaken for event
date.} For Newell's 2013 divestiture completion (gold anchor
2013-07-15), a system cites a gold-attesting retrospective published
2013-09-12 and dates the event to the article, $59$ days off.

\paragraph{Incomplete enumeration.} Partial lists
with no misread signal. For deployable paradigms this bucket coincides with the budget-bound coverage tail quantified by the
top-$k$ sweep (Table~\ref{tab:coverage_at_k}); for the LC-oracle, which holds all gold articles in context, it instead reflects generation-side enumeration limits---the under-enumeration quantified in Table~\ref{tab:setsize}.

\end{document}